\documentclass[twoside]{article}

\usepackage{tech_report}

\usepackage[round]{natbib}

\usepackage{amsmath,amsfonts,bm}

\def\eqref#1{equation~\ref{#1}}
\def\1{\bm{1}}

\DeclareMathAlphabet{\mathsfit}{\encodingdefault}{\sfdefault}{m}{sl}
\SetMathAlphabet{\mathsfit}{bold}{\encodingdefault}{\sfdefault}{bx}{n}

\usepackage{framed}
\usepackage{hyperref}
\usepackage{url}
\usepackage[textsize=footnotesize,color=green!40]{todonotes}
\usepackage{tikz}
\usetikzlibrary{calc,backgrounds}
\usetikzlibrary{arrows,calc,decorations.pathreplacing,positioning,shapes,shapes.geometric,shapes.callouts}
\usetikzlibrary{shapes,intersections,external}
\usepackage{pgfplots}

\newcommand{\logic}{{\mathcal{L}}}
\newcommand{\encode}{{\texttt{encode}}}
\newcommand{\decode}{{\texttt{decode}}}

\definecolor{hublue} {RGB}{  0, 55,108}
\definecolor{hured}  {RGB}{138, 15, 20}
\definecolor{hugreen}{RGB}{  0, 87, 44}
\definecolor{husand} {RGB}{210,192,103}
\definecolor{hugraygreen}{RGB}{209,209,194}
\definecolor{hugrayblue} {RGB}{189,202,211}

\definecolor{distinct11}{HTML}{77aadd}
\definecolor{distinct12}{HTML}{77cccc}
\definecolor{distinct13}{HTML}{88ccaa}
\definecolor{distinct14}{HTML}{dddd77}
\definecolor{distinct15}{HTML}{ddaa77}
\definecolor{distinct16}{HTML}{dd7788}
\definecolor{distinct17}{HTML}{cc99bb}

\definecolor{distinct21}{HTML}{4477aa}
\definecolor{distinct22}{HTML}{44aaaa}
\definecolor{distinct23}{HTML}{44aa77}
\definecolor{distinct24}{HTML}{aaaa44}
\definecolor{distinct25}{HTML}{aa7744}
\definecolor{distinct26}{HTML}{aa4455}
\definecolor{distinct27}{HTML}{aa4488}

\definecolor{distinct31}{HTML}{114477}
\definecolor{distinct32}{HTML}{117777}
\definecolor{distinct33}{HTML}{117744}
\definecolor{distinct34}{HTML}{777711}
\definecolor{distinct35}{HTML}{774411}
\definecolor{distinct36}{HTML}{771122}
\definecolor{distinct37}{HTML}{771155}

\tikzset{
  roundbox/.style={
    rectangle, draw, rounded corners
  },
  arrow/.style={
    ->, >=stealth, rounded corners
  },
  midcircle/.style={
    midway, circle, draw, fill=white
  },
  box p1/.style={
    roundbox,
    fill=white
  },
  box p2/.style={
    roundbox,
    fill=white
  },
  box mpc/.style={
    roundbox,
    fill=distinct12!66
  }
}

\makeatletter
\newcommand{\printfnsymbol}[1]{%
  \textsuperscript{\@fnsymbol{#1}}%
}
\makeatother

\begin{document}

\twocolumn[

\aistatstitle{Data Generation for Neural Programming by Example}

\aistatsauthor{Judith Clymo\printfnsymbol{1} \And
Haik Manukian\printfnsymbol{1}  \And
Nathana{\"e}l Fijalkow \And
Adri{\`a} Gasc{\'o}n \And 
Brooks Paige}

\aistatsaddress{University of Leeds \\
scjc@leeds.ac.uk 
\And 
University of California\\ 
at San Diego \\
hmanukia@ucsd.edu \And
CNRS, LaBRI \\
Alan Turing Institute \\
nathanael.fijalkow@labri.fr
\And
Google \\
adriagascon@gmail.com
\And
UCL \\
Alan Turing Institute \\
bpaige@turing.ac.uk} 
]

\printfnsymbol{1} equal contribution

\begin{abstract}
  {\em Programming by example} is the problem of synthesizing a program from a small set of input / output pairs.
  Recent works applying machine learning methods to this task show promise, but
  are typically reliant on generating synthetic examples for training.
  A particular challenge lies in generating meaningful sets of inputs and outputs, which 
  well-characterize a given program and accurately demonstrate its behavior.
  Where examples used for testing are generated by the same method as training data then the performance of a model may be partly reliant on this similarity.
  In this paper we introduce a novel approach using an SMT solver to synthesize inputs 
  which cover a diverse set of behaviors for a given program.
  We carry out a case study comparing this method to existing synthetic data generation procedures in the literature, and 
  find that data generated using our approach improves both the discriminatory power of example sets and the ability of trained machine learning models to generalize to unfamiliar data.
\end{abstract}

%% 2012 ACM Computing Classification System (CSS) concepts
%% Generate at 'http://dl.acm.org/ccs/ccs.cfm'.
%\begin{CCSXML}
%<ccs2012>
%<concept>
%<concept_id>10010147.10010257.10010293</concept_id>
%<concept_desc>Computing methodologies~Machine learning approaches</concept_desc>
%<concept_significance>500</concept_significance>
%</concept>
%<concept>
%<concept_id>10010147.10010257.10010293.10010294</concept_id>
%<concept_desc>Computing methodologies~Neural networks</concept_desc>
%<concept_significance>300</concept_significance>
%</concept>
%<concept>
%<concept_id>10011007.10011006.10011050.10011056</concept_id>
%<concept_desc>Software and its engineering~Programming by example</concept_desc>
%<concept_significance>500</concept_significance>
%</concept>
%</ccs2012>
%\end{CCSXML}

%\ccsdesc[500]{Computing methodologies~Machine learning approaches}
%\ccsdesc[300]{Computing methodologies~Neural networks}
%\ccsdesc[500]{Software and its engineering~Programming by example}%% End of generated code

%% Keywords
%% comma separated list
%\keywords{Programming by example, program synthesis, data generation}  

\section{Introduction}
% !TEX root =  main.tex

The machine learning community has become increasingly interested in tackling the problem of programming by example (PBE), where the goal is to find a short computer program which is consistent with a set of input and output (I/O) pairs.
Methods have been developed that use neural networks either to generate source code directly \citep{devlin2017robustfill,parisotto2016neuro,bunel2018leveraging} or to aid existing search algorithms \citep{balog2017deepcoder}.
These papers report impressive empirical performance. 
However, deep learning methods are data-hungry, and programming by example is not a data-rich regime.

Many neural program synthesis methods are developed targeting domain-specific languages (DSLs), such as the FlashFill environment \citep{gulwani2012spreadsheet,devlin2017robustfill}, a custom list processing language \citep{balog2017deepcoder,feng2018program}, or the Karel the Robot environment \citep{bunel2018leveraging}.
These languages do not have a large body of human-generated code, nor do they have canonical examples of I/O pairs which correspond to usage in a real-world scenario.
As a result, neural program synthesis has turned to generating synthetic data, typically by sampling random programs from some predefined distribution, sampling random inputs, and then evaluating the programs on these inputs to obtain I/O pairs.

The synthetic data is typically used as both the training set for fitting the model, and the testing set for evaluating the model, which is problematic if there is a mismatch between these random examples and potential real-world usage.
In the absence of any ground truth this problem of over-fitting is often ignored.
In addition, I/O pairs produced by random generation may not be examples that particularly well-characterise a given program. 
This can affect experimental results through {\em program aliasing}, where many different possible programs are consistent with the I/O examples \citep{bunel2018leveraging}.

Nearly all existing approaches are based on an implicit assumption that programs and I/O examples used for training should be chosen in a way that is as uniform as possible over the space,
despite the fact that many I/O examples are uninformative or redundant, as is well known in the automated software testing community (see e.g.\ \citet{godefroid2005dart}).

\paragraph{Contributions.}
In this paper we consider how to generate sets of I/O examples for training a neural PBE system.
In particular we are concerned with the generalizability of neural networks trained for PBE on synthetic data.
We present four approaches to data generation in this context, including a novel constraint based method.
We show how these different methods can be applied in a case study of the DeepCoder DSL from \citet{balog2017deepcoder}, an expressive language for manipulating lists of integers.
The DeepCoder DSL has several features that make it an interesting and challenging setting for generating synthetic training data: it is capable of displaying complex behavior including branching and looping; the set of valid inputs for a given program can be difficult to define and/or restricted; and the most informative examples are distributed unevenly throughout the space.
We train neural networks designed for program generation and assisting a program search using examples produced by the four outlined methods, then use cross-comparison to quantify the robustness of these networks. 
We also evaluate the degree to which different synthetic data sets uniquely characterise a program, important for addressing the program aliasing problem.

\section{Related work}

%\paragraph{How do existing approaches generate data?}
%In order for neural program synthesis methods to have sufficient training data, it is often necessary to augment any real-world examples with machine generated programs and input-output pairs.
Most approaches to generating I/O examples are based on random sampling schemes, with either a rejection step or initial constraints.

\citet{balog2017deepcoder} construct a database of programs in the DeepCoder DSL by enumerating programs up to a fixed length and pruning those with obvious problems such as unused intermediate values.
They produce I/O examples for each program by restricting the domain of inputs to a small, program-dependent subset which is guaranteed to yield valid outputs and then sampling uniformly from this set.
\citet{feng2018program} targets the same DSL but example pairs are generated by sampling random inputs and seeing whether they evaluate to a valid output. 
If a sufficient number of valid pairs are not found within a fixed amount of time, the program is discarded.
The paper does not state what distribution is used to sample inputs.
In both of these papers, this synthetic data is used for both training the model and for evaluating its performance.

\citet{bunel2018leveraging} considers the Karel the Robot domain, and creates a dataset of programs by random sampling from the production rules of the DSL, pruning out programs which represent the identity function or contain trivial redundancies.
I/O examples are constructed by sampling and evaluating random input grids.
No mention is made of what sampling distribution is used for either the programs or the input grids;
the synthetic data is used for both training and testing.

\citet{parisotto2016neuro} generates synthetic data for the FlashFill domain \citep{gulwani2012spreadsheet} by sampling random programs up to a maximum of 13 expressions.
The reported performance suggest a very large gap between the synthetic examples (97\% accuracy) and the real-world data  (38\% accuracy).
\citet{devlin2017robustfill} is more cautious: performance is evaluated only on the real-world examples, with synthetic data used only for training.
The training data is produced by simulating random programs up to a maximum of 10 expressions, and then constructing I/O pairs by sampling random inputs and testing to see whether executing the program raises an exception.
Neither paper explicitly specifies any of the sampling distributions. 

The papers above all primarily focus on advancing search algorithms or improving heuristics, with little attention to data generation;
the only exception we are aware of is \citet{shin2019synthetic}, which specifically proposes a method for improving synthetic example generation.
Their approach requires first defining a set of ``salient variables'' for the domain: these take the form of a mapping from a synthetic training example to a discrete value.
They then define a sampling strategy for examples which aims to maximize their entropy, by generating a training dataset which is overall approximately uniform over all the combinations of the different discrete salient variables.
Particular attention is paid to the Karel the Robot domain, and the generation of synthetic data for the algorithms in \citet{bunel2018leveraging}.

When there are many salient variables the discrete domain can be large. 
However, the number of I/O examples needed for each program is typically quite small. 
In the Karel the Robot domain the input space is not constrained by the programs so it is natural to enforce uniformity across salient variables over the whole set of inputs. 
Where the program can significantly affect the valid input space the approach is not appropriate. 
The contributions we make to this task are complementary to those of \citet{shin2019synthetic}, which in this domain could still be beneficially applied in selecting a set of synthetic programs, but would be difficult to adapt to generating meaningful inputs.

\section{ML-aided Programming by Example}
\label{sec:background}
% !TEX root =  main.tex

\begin{figure}
	\resizebox{\columnwidth}{!}{%
		\begin{tikzpicture}
		\node[] (ref) at (5, 0) {};
		
		\node[roundbox, thick, align=center, scale=1, inner sep=4] (ml) at ($(ref.south)+(-4,0)$) {%
			{\textsc{Learning}}\\[1ex]
			{\textsc{Model} M}
		};
		
		\node[align=center] (input) at ($(ml.west)+(-2,0)$) {%
			Examples\\$(x_1, y_1), \ldots, (x_k, y_k)$
		};
		
		\node[draw, shape border rotate=90, aspect=0.10, cylinder, thick, inner sep=4, align=center] (d) at ($(ml.north)+(0, 1.5)$) {%
			\scriptsize {\em Synthetic}\\
			\scriptsize Training\\ \scriptsize Dataset
		};
		
		\node[roundbox, thick, align=center, scale=1, inner sep=4] (gen) at ($(d.east)+(-3.5,0)$) {%
			{\textsc{DSL-specific}}\\
			{\textsc{Data Generation}  G}
		};
		
		\node[draw, regular polygon, regular polygon sides=3, thick, dashed, align=center, font=\tiny] (tree) at ($(ml.east)+(2.5,-.5)$) {%
			Prioritised\\Search Tree
		};
		\path[arrow, line width=1pt] ($(ml.east)+(0, 0)$) edge [] node[align=center, font=\small, yshift=17] {Ranking\\heuristic}  ($(tree.west)+(0, .5)$);
		\path[arrow, line width=1pt] ($(input.east)+(0, 0)$) edge [] node[] {}  ($(ml.west)+(0, 0)$);
		\path[arrow, line width=1pt] ($(d.south)+(0, 0)$) edge [] node[] {}  ($(ml.north)+(0, 0)$);
		\path[arrow, line width=1pt] ($(gen.east)+(0, 0)$) edge [] node[] {}  ($(d.west)+(0, 0)$);
		
		\end{tikzpicture}
	}
	\caption{ML-aided programming by example}
	\label{fig:ml-guided-synt}
\end{figure}
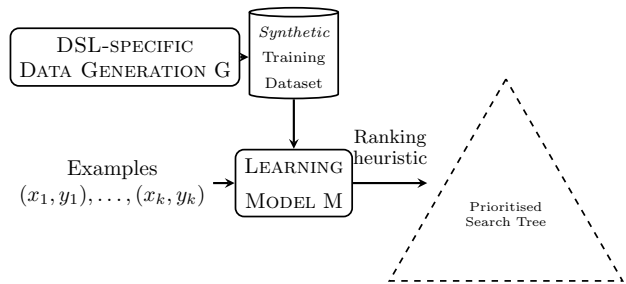

Figure~\ref{fig:ml-guided-synt} outlines the general approach to programming by example we take in this paper, 
providing a framework which can be used to understand methods and settings employed in recent work. 
The problem is defined in terms of a concrete DSL that takes as input a set of examples $(x_1,y_1),\dots,(x_k,y_k)$, and synthesizes a program mapping inputs $x_i$ to the corresponding outputs $y_i$. 
At the core of the approach is a search algorithm that explores the space of all programs up to a certain predefined length according to a ranking function. 
This heuristic ranking function is derived from the output of an ML model $\textsc{M}$ that inspects the input $(x_1,y_1),\dots,(x_k,y_k)$. 
Crucial in this approach is the training dataset of $\textsc{M}$. 
This dataset is obtained using a DSL-specific generation procedure $\textsc{G}$ that generates (a) programs and (b) a small set of inputs for those programs.

Hence, to instantiate a concrete approach to ML-aided programming by example one needs to define (i) the DSL, (ii) the data generation procedure $\textsc{G}$, and (iii) the ML model $\textsc{M}$ and corresponding search procedure. 
As mentioned above, previous work focuses primarily on (iii). 
Here we instead focus on (ii), the training data generation step, and investigate its effect in concrete instances of the complete pipeline. 

\subsection{Domain specific language}
\label{sec:dsl}

\begin{figure*}[t]
\centering
\resizebox{0.95\textwidth}{!}{
\fbox{
\begin{minipage}{.3\textwidth}
\texttt{a} $\gets$ \texttt{[int]}\\
\texttt{b} $\gets$ \textsc{Sort} \texttt{a}\\
\texttt{c} $\gets$ \textsc{Filter} \texttt{(>0)} \texttt{b}\\
\texttt{d} $\gets$ \textsc{Head} \texttt{c}\\
\texttt{e} $\gets$ \textsc{Drop} \texttt{d} \texttt{b}\\
\end{minipage}%
\begin{minipage}{.7\textwidth}
\textbf{An input-output example:}\\
\emph{Input}:\\
\texttt{[-17, -3, 3, 11, 0, -5, -9, 13, 6, 6, -8, 11]}\\
\emph{Output}:\\
\texttt{[-5, -3, 0, 3, 6, 6, 11, 11, 13]}\\
\end{minipage}
}
}
\caption{An example program in the DSL that takes a single integer array as its input.}
\label{fig:met:ExampleLINQ}
\end{figure*}

The DeepCoder DSL \citep{balog2017deepcoder}, which we will use as our running example, consists of high-level functions manipulating list of integers; a program is a sequence of functions which take as input any of the inputs or previously computed values.
The DSL contains the first-order functions
\textsc{Head}, \textsc{Last}, \textsc{Take}, \textsc{Drop}, \textsc{Access}, \textsc{Minimum}, \textsc{Maximum}, 
\textsc{Reverse}, \textsc{Sort}, \textsc{Sum},
and the higher-order functions \textsc{Map}, \textsc{Filter}, \textsc{Count}, \textsc{ZipWith}, \textsc{Scanl1}. 
The higher-order function \textsc{Map} can be combined with \texttt{(+1)}, \texttt{(-1)}, \texttt{(*2)}, \texttt{(/2)}, \texttt{(*(-1))}, \texttt{(**2)}, \texttt{(*3)}, \texttt{(/3)}, \texttt{(*4)}, \texttt{(/4)}.
The two higher-order functions \textsc{Filter} and \textsc{Count} use predicates
\texttt{(>0)}, \texttt{(<0)}, \texttt{(\%2==0)}, \texttt{(\%2==1)}.
Finally, \textsc{ZipWith} and \textsc{Scanl1} are paired with 
\texttt{(+)}, \texttt{(-)}, \texttt{(*)}, \textsc{Min}, \textsc{Max}.
The lists in a program are of a predefined length ($20$ in the experiments of \cite{balog2017deepcoder}) with values in $[-255,256]\cup \{\bot\}$, with $\bot$ denoting an {\em undefined} value.
The semantics of the DSL assume some form of bound checking, as indexing a list out of its bounds returns $\bot$, and over(under)-flow checking, as values outside of the range $[-255,256]$ evaluate to $\bot$. 
As we will see, this impacts data generation by constraining the domain of valid inputs.
This DSL is remarkably expressive: it can express non-linear functions via repeated integer multiplication, control flow (if-then-else statements can be encoded by means of \textsc{Filter}), and loops over lists (by means of the higher-order functions). 
Moreover, the higher-order functions like \textsc{Reverse}, \textsc{Sort}, \textsc{ZipWith} and \textsc{Scanl1} allow to encode surprisingly complex procedures in just a few lines, such as the four-line example shown in Figure~\ref{fig:met:ExampleLINQ}.

\section{Data Generation Methods}
% !TEX root =  main.tex

We propose three new approaches for the data generation step.
The first is based on a probabilistic sampling of the input space, while the  other two treat data generation as a constraint satisfaction problem.  We develop a general framework for this approach and show how additional constraints can be used to impose variation in the I/O pairs produced. 
 
\subsection{Sampling-based approaches}

We use the approach from~\citet{balog2017deepcoder} as a baseline.
Programs are evaluated in reverse to derive a `safe' range for input values that is guaranteed to produce outputs in a target range.
Values are then sampled uniformly from the safe input range to create the input(s), and the output is calculated by evaluating the program.  
If the safe range for some input is empty or a singleton, then the program is discarded.
In reporting results of our experiments, we refer to this method of data generation as ``Restricted Domain''.

The purpose of the reverse propagation of bounds is to exclude all non-valid inputs from the sampling.
However, some valid inputs are also excluded.
Instead, inputs can be sampled from an over-approximation of the set of valid inputs or by using a probability distribution that prioritises parts of the input space where valid inputs are known to be common.
Inputs are then rejected if they are unsuitable for the program being considered.

We observe that in the DeepCoder DSL, when the output range is bounded, small input values are compatible with more programs than large values.
Sampling with a bias towards small values means that suitable inputs are found with high probability for all programs within a fixed number of attempts $n$. 
If suitable inputs for a program remain common outside of the safe range identified by back propagation of bounds then the gain from allowing these to be included could be significant.

To produce a sample, we first fix a random length of the input (uniformly at random) and generate a value $r$ from an exponential distribution \mbox{i.e.} with probability $P(r) \sim \text{Exp}(r, \lambda) = \lambda e^{-\lambda r}$.
The input values are selected uniformly at random (with replacement) from the range $[-r, r]$, then evaluated on the program. 
If the output of the program is within the desired range the pair is accepted, otherwise a new value of $r$ is sampled and the process repeated.

Varying the choice of $\lambda$ modifies the strength of the bias, affecting both the number of attempts that must be allowed and also the similarity of examples generated. 
In our case, we chose $\lambda = 0.001$ and $n = 500$.
This method will be referred to as ``Non-Uniform Sampling'' in the experiments section.

\subsection{Constraint-solving based approaches}

Sampling methods are not well suited to navigating valid input spaces that are sparse or irregular, and in a language capable of displaying complex behavior the most informative examples may also be rare.
A constraint solver is able to find suitable inputs in parts of the input space that are almost always excluded in the sampling methods. In this section we present a methodology to synthesize a set of examples that relies on constraint solving.

\begin{figure}
\resizebox{\columnwidth}{!}{%
\begin{tikzpicture}

  \node[] (u3) at (5, 0) {};

     \node[align=center] (p) at ($(u3.south)+(-3,0)$) {%
       Program $P(X,Y)$\\
       Logic $\logic$
     };
  
     \node[roundbox, draw, dashed, thick, fill=white, inner sep=5, minimum size=1cm, align=center] (gen) at ($(u3.south)+(0, -2)$) {%
       ~~\\[1ex]
       $\phi(X,Y) = \texttt{encode}_{\logic}(P)$\\[1ex]
       $\psi_i(X,Y) = \texttt{SynCons}_{\logic}(P, \{x_i, y_i\}_i)$\\[1ex]
       $\rho_i(X,Y) = \texttt{SemCons}_{\logic}(P, \{x_i, y_i\}_i)$\\[1ex]
    };

    \node[circle, draw, fill=distinct11!66] (solver) at ($(gen)+(5,0)$) {%
        {\scriptsize $\logic$-\textsc{Solver}}%$S$
    };

    \node[fill=white] (gentext) at ($(gen.north)+(0,0)$) {\textsc{Constraint Controller}};

    \path[arrow, thick] ($(gen.north)-(0, -.1)$) edge [bend left] node[above, align=center] {$\bigwedge_i \psi_i\wedge \rho_i\wedge\phi$} ($(solver.north west)$);
    \path[arrow, thick] ($(solver.south west)$) edge [bend left] node[below,align=center] {$(X\mapsto x_i, Y\mapsto y_i)$} ($(gen.south)-(0, .01)$);
     \path[arrow, thick] ($(p.south)+(.1, .15)$) edge [] node[] {}  ($(gen.north west)+(.5, .05)$);

  \end{tikzpicture}
}
\caption{Our constraint-solving based approaches are instances of the above model.
A constraint controller adaptively produces problems
for the solver, whose solutions correspond to new input-output pairs to be added to the
training dataset}
  \label{fig:solve-based-method}
\end{figure}
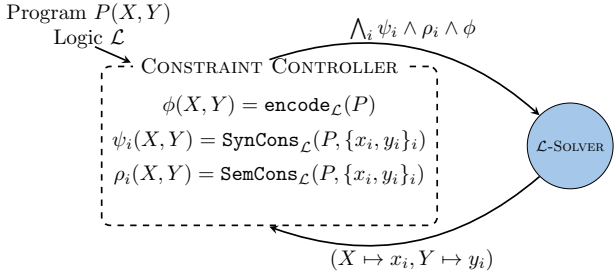

The general procedure is presented in Figure~\ref{fig:solve-based-method} and consists of a
feedback loop between a constraint controller and a solver. The former adaptively produces problems
for the solver, whose solutions correspond to new I/O pairs to be added to the
training dataset. The approach is parametrised by a choice of a logic $\logic$,
and assumes a solver, denoted $\logic$-Solver, that can decide $\logic$, i.e. find
values for the variables in any formula $\phi$ from $\logic$ for $\phi$ to evaluate to true,
or report ``unsatisfiable'' if such values don't exist.
We require encoding and decoding procedures $\encode_\logic, \decode_\logic$ such that
$\encode_\logic$ takes a program $P$ with input variables $X$ and output variables $Y$, and produces a formula whose satisfying assignments
can be translated by $\decode$ into concrete input-output pairs of the program $P$.
For simplicity, we assume that the domain of the variables in $\phi_\logic$
and $P$ is the same in Figure~\ref{fig:solve-based-method} and omit the decoding step.

The simplest data synthesis procedure consists of the controller simply calling
the solver iteratively to collect a
sequence of input output pairs $(x_1,y_1), \ldots, (x_n, y_n)$ for $P$.
Note that for DSLs like the one presented above, a complete decision procedure might
be too much to ask, but for our purposes soundness is sufficient. More concretely,
$\encode_\logic$ should be constructed so that, for each program $P$
in the DSL the following holds:

$$\forall XY:\encode_{\logic}(P) \Rightarrow P(X) = Y$$

This procedure, as described so far, is not very helpful, as nothing
prevents the solver from returning the same pair $(x, y)$
at every iteration. This can be easily avoided by the controller imposing the additional constraint
$X\neq x_i \vee Y\neq y_i$ after every iteration. This is the most basic
form of additional constraint that we consider.
More generally, our approach considers two additional sets of constraints,
which we call {\em syntactic constraints}, denoted
$\texttt{SynCons}_{\logic}$ and {\em semantic constraints},
denoted $\texttt{SemCons}_{\logic}$. While both of this constraints
are conditions on I/O pairs -- either for each line of the program or the whole program --
the former consist of simple equality and inequality checks such as ``input and output
should be different''. In contrast,
semantic constraints are more powerful -- and costly to enforce -- as they
enforce predicates on input output pairs such as ``the maximum value of
the input should be smaller than the maximum value of the output''.

The next two concrete approaches to data generation are instances of the above scheme, and thus correspond to concrete
choices for $\logic$, $\encode$, $\decode$, $\texttt{SynCons}_{\logic}$, and $\texttt{SemCons}_{\logic}$, and the corresponding
variations in the behavior of the constraint controller. In our implementation we use Z3~\citep{z3} as a back-end solver.
Z3 implements incremental solving, which allows to push and pop constraints into an existing formula while
preserving intermediate states, a crucial aspect of the implementation of a controller.
As per $\logic$, we experimented with the theory of non-linear arithmetic, for which Z3 implements incomplete procedures,
and the theory of bitvectors.

\subsubsection{Simple constraint solving}
Our third data generation method uses a constraint solver to produce valid examples with only minimal extra guidance. Hence, in this case $\texttt{SemCons}_{\logic}(P, S) = \emptyset$.
$\texttt{SynCons}_{\logic}$ are given in Table~\ref{tab:relations} by means of a list of constraints $c_1,\ldots,c_5$. At the $i$th iteration the pair $(x_i,y_i)$ is obtained from the solver as a satisfying assignment of the constraints $\bigwedge_i \psi_i\wedge \phi$ (as shown in Figure~\ref{fig:solve-based-method}), where $\psi_i = \bigwedge_{i=1}^{5} c_i$. This guarantees that synthesized examples will satisfy these conditions if possible.

\begin{table}[tp]
	\caption{Constraints and predicates used in Syntactic and Semantic constraints used in our constraint-based input generation algorithms. Constraints and predicates are depicted as conditions on the I/O pairs $(x_1, y_1), \ldots, (x_5, y_5$) generated by the constraint-based approaches. For programs taking more than one input all constraints are applied to each of the inputs. $x_i^l$ is the intermediate program output of line $l$ given input $x_i$.
		}
	\begin{center}
        \resizebox{\columnwidth}{!}{%
		\begin{tabular}{l}
			\hline
                        
			$\texttt{SynCons}_{\logic}$ for ``Constraint Based'' and ``Semantic Variation''\\
			\hline
			$c1 := \forall i: x_i\neq y_i$ \hfill {\small\em Output does not match input.} \\
			$c2 := \forall i:x_i \neq [~]$ \hfill {\small\em All inputs are not empty.} \\
			$c3 := \forall i\neq j: x_i\neq x_j$ \hfill {\small\em No duplicate inputs in a set.} \\
			$c4 := \forall i\neq j: y_i\neq y_j$ \hfill {\small\em No duplicate outputs in a set.} \\
			$c5 := \forall i: |x_i| > B$ \hfill{\small\em Input length larger than random bound $B$.}\\
			\hline\\[1ex] \hline 
			Predicates in $\texttt{SemCons}_{\logic}$ for ``Semantic Variation''\\
			\hline
			$p_1 := \max(y_i) > \max(y_{i-1})$\hfill {\small\em Increase maximum.}\\
			$p_2 := \min(y_i) < \min(y_{i-1})$ \hfill{\small \em Decrease minimum.}\\
			$p_3 := |x_i| > C$ \hfill{\small\em  Input length larger than $C$.}\\
			$p_{l,4} := head(x_i^l) \neq head(y_i^{l-1})$ \hfill{\small\em Change in head.}\\
			$p_{l,5} := last(x_i^l) \neq last(y_i^{l-1})$ \hfill{\small\em Change in last.}\\
			$p_{l,6} := |x_i^l| \neq |y_i^{l-1}|$ \hfill{\small\em Change in length.}\\
			$p_{l,7} := max(x_i^l) \neq max(y_i^{l-1})$ \hfill{\small\em Change in maximum.}\\
			$p_{l,8} := min(x_i^l) \neq min(y_i^{l-1})$ \hfill{\small\em Change in minimum.}\\
                 
			\hline
		\end{tabular}
                }
	\end{center}
	\label{tab:relations}
\end{table}

\subsubsection{Constraint solving to generate varied examples}
Our fourth method  is inspired by program verification approaches such as predicate abstraction and CounterExample-Guided Abstraction Refinement loop (CEGAR)~\citep{ClarkeGJLV03}. This method uses constraints more aggressively, and in particular semantic constraints, by implementing an adaptive constraint controller. The syntactic constraints used in this case are as in the previous method, so we focus on describing the semantic constraints, and the adaptive behaviour of the controller.

The controller first uses sampling to generate a small set of examples, and then keeps track of the evaluation of predicates $p_1, \ldots,p_3$ for the output of the program and $p_4, \ldots,p_8$, {\em for every line of the program} (see Table~\ref{tab:relations}) for each of the inputs found so far. For example, for $5$-line programs this corresponds to storing a $5\times 5$ Boolean matrix $M^k$ for each input $x_k$ so far. The controller also maintains a record of constraint combinations that result in unsatisfiable problems so these can be avoided in subsequent calls. The high-level idea is as follows: 
firstly, by monitoring any bunching of in previous examples the controller can recognise programs with valid examples tending to bunch in one part of the I/O space (in this DSL, bunching is always towards zero, so monitoring maximum and minimum is a good proxy) and force the constraint solver to seek out examples in parts of the space where valid examples are relatively rare; 
secondly by monitoring which of the predefined behaviours $p_4,\ldots,p_8$ are satisfied for each program line, by each of the inputs $x_i$ collected so far, the controller can detect sets inputs that do not sufficiently exercise internal lines of the program, and hence send a weak signal regarding the presence of the function at that line to the output. 
Besides recording which behaviors are exhibited by each line, the controller can impose a given behavior by asserting a specific predicate (or its negation) as part of the semantic constraints for that iteration. More specifically, the controller defines the semantic constraint of iteration $i$ as $\texttt{SemCons}_{\logic}(X,Y) := (\forall k < i:\exists n, m: (p_{n, m}(X, Y) \neq M^k_{n,m}))$, where $p_{n,m}$ is a Boolean formula encoding that the $n$th line of the program satisfies the $m$th predicate on input $X$. This constraint enforces that in any valid assignment $X\mapsto x_i, Y\mapsto y_i$ obtained by the solver, input $x_i$ will induce a behavior that differs with each combination of behaviors induced by inputs found so far in at least one line of the program. This makes the intuitive goal of finding ``a varied set of inputs'' precise.
\paragraph{Choosing predicates.}
The choice of features to monitor is specific to the DSL and based on abstractions of the actions of its constituent functions.
In this DSL we are working with lists of integers defined by their length, the set of values contained, and the order in which those values appear. The functions of the DSL can change all of these features.
Our aim is that if the program is capable of changing the values that appear in the output compared to the input then there should be an I/O example that demonstrates this, if it is capable of reordering the elements this should also be shown, and so on. 
The maximum, minimum, first, and last elements of a list act as simplified indicators for changes in the values and order of the list and are the same features used in \citet{feng2018program} to assist the search procedure for this DSL. 

We have referred to this data generation method as ``Semantic Variation'' in the experiments.

\section{Experiments}
% !TEX root =  main.tex

\begin{figure}[t]
	\centering
	\includegraphics[width = 0.425\textwidth]{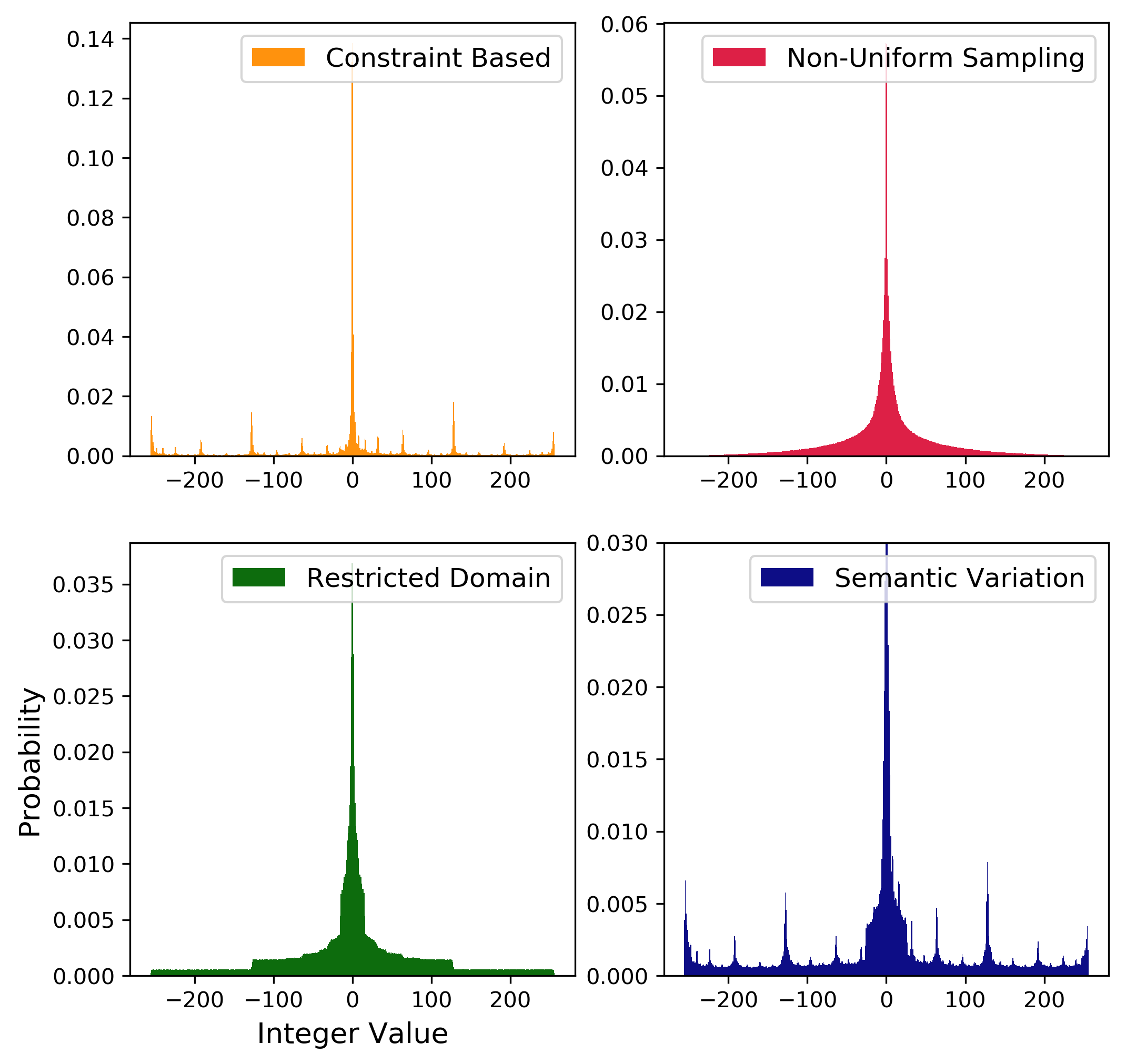}
	\caption{Histograms of the input values found in training sets generated with the four data generation methods considered. Note the large variation in y-axes.
	\vspace{-0.2cm}
	}
	\label{fig:inherent-dsl-bias}
\end{figure}

We consider two choices of ranking heuristics; one exactly following~\citet{balog2017deepcoder}, and the other a natural extension based on recurrent neural networks, following a sequential generation paradigm~\citep{devlin2017robustfill,bunel2018leveraging}.

The four synthetic data generation methods discussed above are compared in each of these settings.
In particular, we investigate how neural networks trained with data from one method perform at test time on data generated by another. 
For a fair comparison of the data generation methods, we use the same split of programs into training and testing examples.
Figure~\ref{fig:inherent-dsl-bias} shows how input values are distributed in data generated by each method.

\paragraph{The DeepCoder heuristic.}
The ranking function used in \citet{balog2017deepcoder} 
estimates, for each of the 38 functions in the DSL, the conditional probability that the function ever appears in the program.
The predicted probabilities provide a ranking for a depth-first search.

\paragraph{The recurrent neural network heuristic.}
In addition, we consider an extension to DeepCoder which is loosely inspired by the recurrent neural network architectures used for program generation in other DSLs \citep{devlin2017robustfill,bunel2018leveraging}.
Instead of training the network to output a single set of probabilities, we train a network to output a sequence of probabilities, conditioned on the current line of the program.

We do this by modeling the sequence of lines with a long short-term memory (LSTM) network \citep{hochreiter1997long}.
To ease comparison, we leave the majority of the network architecture unchanged;
the only modification is the penultimate layer is replaced by an LSTM.
The result is a network which takes as arguments not just the I/O examples, but also a target ``number of lines'', and then returns estimates of probabilities that a function occurs on a per-line basis, rather than program-wide.
Complete architectural details for both networks are in the appendix.

\begin{figure}[t]
	\centerline{\includegraphics[width = \columnwidth]{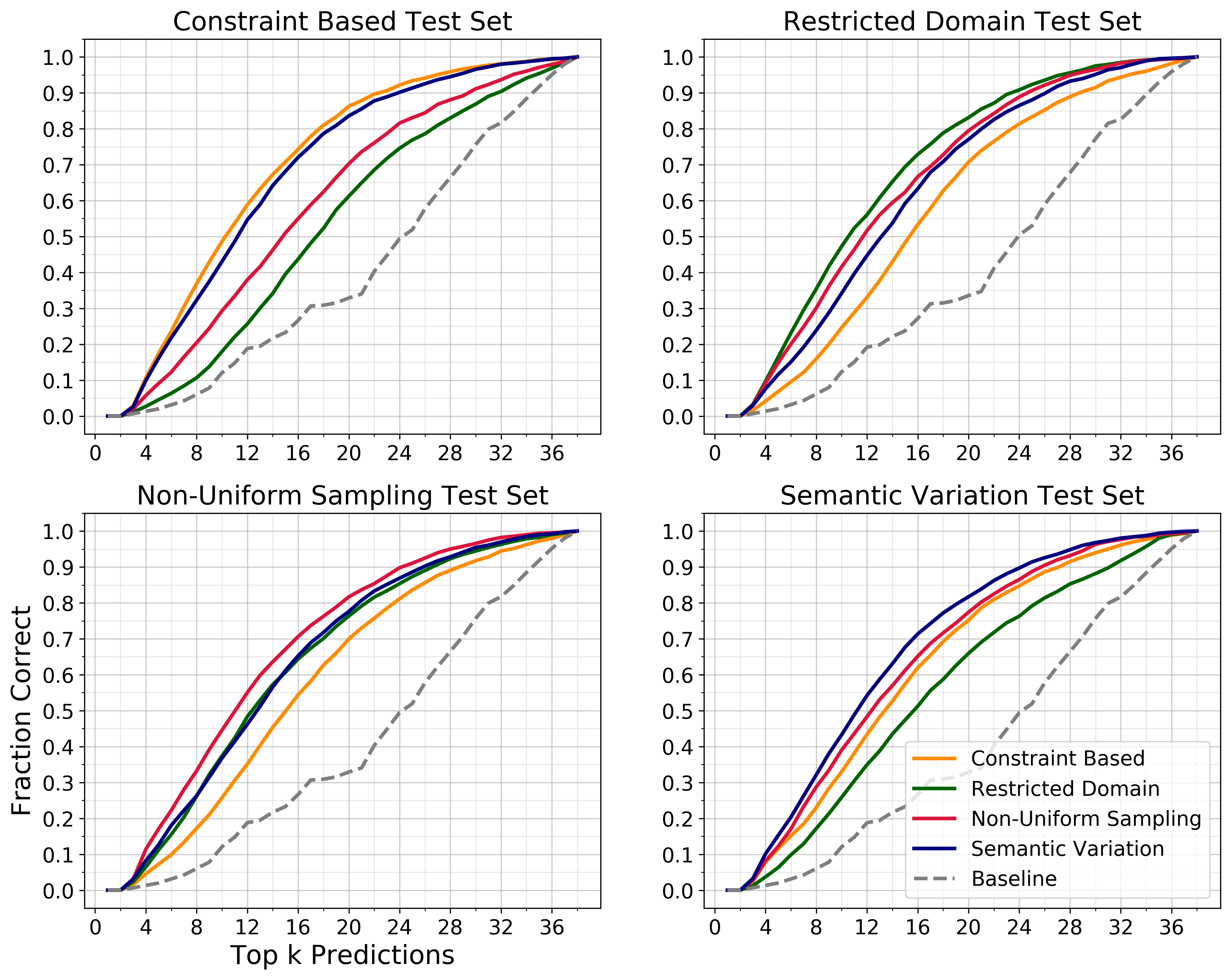}}
	\caption{Each network predicts the functions in a given test set program. We report the fraction of programs for which all constituent functions are contained in the top-$k$ predictions. The networks are trained on one mode of generation and tested on all the others. Each line represents the performance of a training set, defined in the legend. 
		From the performance gaps, we see that the network trained on Semantic Variation data appears more robust to a change in test data generation method.
	}
	\label{fig:top-k-plot}
\end{figure}

\subsection{Cross-generalization of different methods}
The plots in Figure~\ref{fig:top-k-plot} show the proportion of programs for which every line is included within the top $k$ predictions given by a neural network. 
If functions actually present in the program are ranked highly by the network, this will accelerate the runtime of any corresponding guided search. 
The baseline is given by a fixed ordering reflecting the relative frequency of each function in the set of programs used for training.
All networks performed better than the baseline on all test sets, showing that the network is able to generalise beyond its training setting.
Test data is always most accurately interpreted by the network trained on data generated through the same process;
in fact, the four networks performed almost identically to each other when tested on their own data.
However, the ability of each network to transfer to foreign data, from a different generation process, varied significantly. 
The two sampling methods (Restricted Domain, Non-Uniform Sampling) in particular found the data generated by an SMT solver difficult, while the loss from the constraint-based methods to the sampling data was smaller, though still marked.

The sum of the area under the curves indicates the robustness of the networks, with the Semantic Variation method having the largest area; values shown in Table~\ref{tab:AUCs}.

\begin{table}[t]
	\caption{Total area under top-$k$ curves, across all approaches to test data generation.}
	\begin{center}
		\begin{tabular}{|c|c|}
			\hline
			Method & Total AUC \\
			\hline
			Semantic Variation & 126.11 \\
			Non-uniform Sampling & 123.14 \\
			Constraint Based & 118.24 \\
			Restricted Domain & 114.21 \\
			\hline
		\end{tabular}
			\vspace{-0.2cm}
	\end{center}
	\label{tab:AUCs}
\end{table}

\subsection{Program Aliasing}
\begin{table}[tp]
	\caption{Errors due to program aliasing. \textit{Error} corresponds to programs which are correct on the test data, but differ from the original generating program; the predicted programs tend to be \textit{shorter}, and often strictly \textit{contained} within the target program.}
	\label{tab:error}
	\begin{center}
	        	\resizebox{\columnwidth}{!}{
		\begin{tabular}{|c|c|c|c|}
			\hline
			Method & \% Error & \% Shorter & \% Contained \\
			\hline
			Semantic Variation & 4 & 2 & 2\\
			Constraint Based & 6 & 4 & 3\\
			Non-uniform Sampling & 6 & 5 & 4\\
			Restricted Domain & 15 & 12 & 9\\
			\hline
		\end{tabular}}
	\end{center}
\end{table}

We measure how well the generated example sets characterise their target program by taking a sample of programs and I/O sets generated by each method, and searching for programs no longer than the intended program which matched the given examples. 
If an alternative program of the same length was recorded across all test sets then this could be due to logical equivalence, and these examples were excluded.

The Restricted Domain method had the highest rate of ambiguous example sets with $15\%$ of the sets tested able to be satisfied by an alternative (not equivalent) program, compared to a $4\%$ error rate on examples generated by the Semantic Variation method.
Some errors were very subtle, confusing programs that were logically different only on inputs with a specific form.
Others were due to difficulty distinguishing the functions that output integers: in a set of only five examples it is relatively common that the maximum of every list is also at the same position, for example.

The results are summarised in Table~\ref{tab:error} which shows the percentage of example sets where an alternative program was found and also indicates whether this alternative program was strictly shorter and strictly contained within the target program (i.e.\ the target program included some function had no discernible effect for any of the provided inputs).
\subsection{Performance on Independent Test Sets}  

\begin{figure*}[t]
	\centerline{\includegraphics[width = 0.78\textwidth]{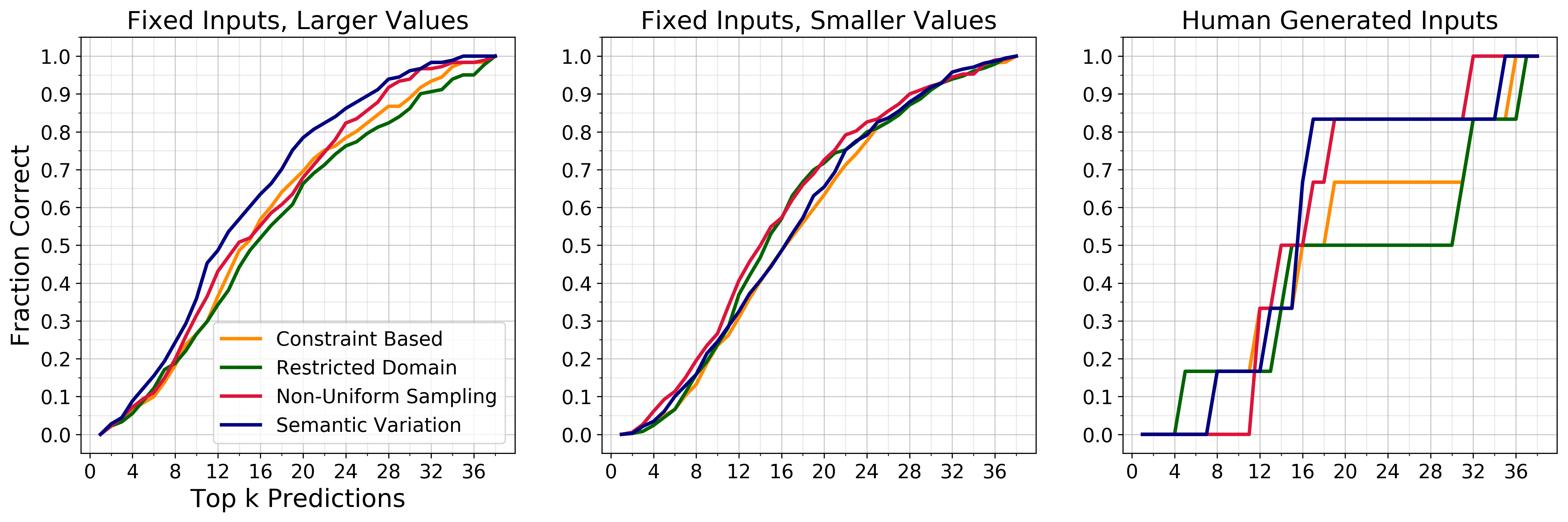}}
	\vspace{-1em}
	\caption{The fraction of programs contained in the top-k predictions of trained networks on hand-crafted examples. Left and middle figures show performance on problems sets with fixed inputs with large and small average values respectively. Right figure shows performance on a small set of human generated examples.}
	\label{fig:human}
\end{figure*}

\begin{figure}[t]
	\centering
	\includegraphics[width = 0.435\textwidth]{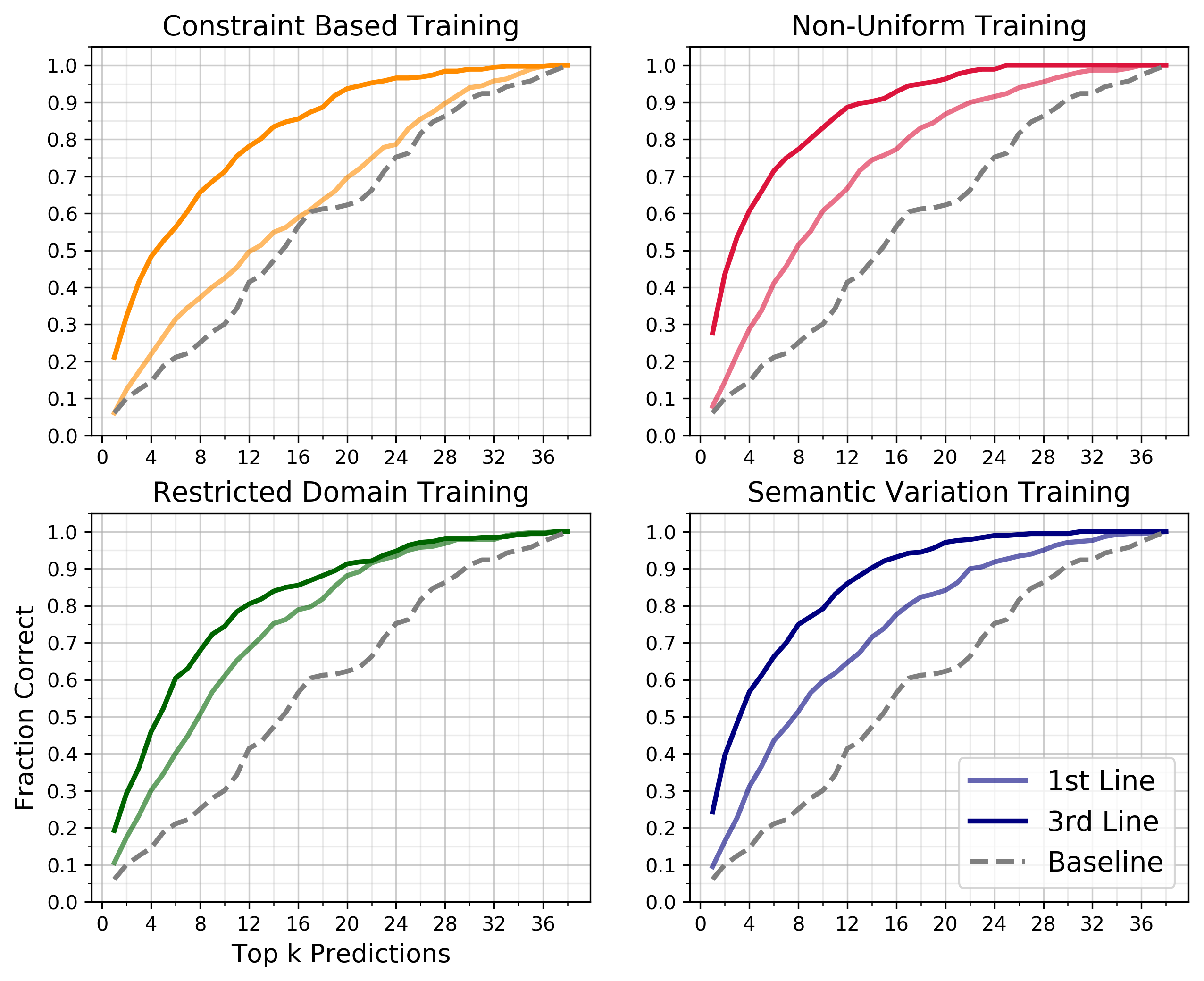}
	\caption{First and last line predictive accuracy of the recurrent architecture on our test set with fixed inputs.}
	\label{fig:top-k-rnn}
\end{figure}

We sought to test the four networks on data that is not generated by a machine and created a small hand-crafted test set of examples on programs of length two to five. 
The performance of the neural networks is shown in Figure~\ref{fig:human}, in the right-most plot.
The test set is clearly too small to draw any serious conclusions, but the good performance of constraint based approaches is encouraging.

As an alternative to this, we fixed five inputs 
and ran a set of programs on these same five inputs, keeping those for which the resulting example set could not be satisfied by another program of the same or smaller length.
Because some functions require mainly small values in the inputs we made one test set by this method which had almost all input values between $ -10 $ and $10$, and a second set with many more large values. 
This approach ensures the inputs alone are relatively uninformative, forcing the networks to derive their predictions from the relationship between the inputs and outputs. 
The results of this experiment are also shown in Figure~\ref{fig:human} (left and centre), again giving the proportion of example sets for which the whole program was contained in the top $k$ predictions. 

The networks all performed better than random, showing their predictions are not purely reliant on learning something about how the examples are generated.
We also observe a difference in performance depending on whether the examples contain mostly small or mostly large values. 

Separately, we evaluate the performance of the recurrent neural network architecture, detailed in the appendix, over this difficult set of examples. The same RNN model was trained over sets generated with the four different methods, as for the feed-forward networks. We plot the fraction of times the RNN was correct about the line appearing in the top-$k$ predictions for that line in Figure~\ref{fig:top-k-rnn}. 
We see that the last line is easier to predict than the first, for all data generation methods, matching the intuition that as more functions are applied to the data, more information is lost about lines that occurred earlier in the program. Overall, when compared to baseline, some sets struggle more than others, but much like the feed-forward case, there is no strategy that has a commanding edge over the others. % the test set. 

However, in longer programs than considered in this paper, we imagine that the RNN architecture could more substantially assist a search than the feed-forward networks. 
The network considered here provides an ordering for all lines at the start of the search.
A more advanced approach could update its prediction for the next line given the evolving state of a partial program;
we leave this to future work as it requires careful consideration of runtime costs:
Repeated re-evaluation of an RNN inside the inner loop of the search algorithm could be inefficient relative to simply running more iterations with a cheaper heuristic, whereas the per-line ranking used here has identical search-time cost as the DeepCoder heuristic.

\section{Discussion}

All the methods of generating data that we have considered were useful in training the neural network, and each of the four trained networks displayed the ability to generalise to unfamiliar testing data. 
Conversely, all networks displayed a preference for testing sets generated in the same way as the data used for training, demonstrating over-fitting to the data generation method to some degree.

Constraint solving proved an effective way of discovering examples that were out of reach to sampling methods, and simple local features were useful in creating example sets that characterise the target program well and reduce the occurrence of ambiguous examples. 
We also saw that the more informative training data in the Semantic Variation method increased the resilience of the neural network against unfamiliar test scenarios.

\section*{Acknowledgments}
\begin{small}
Work started when J. C and H.M were at the The Alan Turing Institute for the Summer 2018 Internship Programme,
N. F and A. G. were at the The Alan Turing Institute and Warwick University, and 
B. P at the The Alan Turing Institute and UCL.
All were supported by The Alan Turing Institute under the EPSRC grant EP/N510129/1, 
and the UK Government's Defence \& Security Programme in support of the Alan Turing Institute.
\end{small}

\bibliography{bib}

\clearpage
\appendix
% !TEX root =  main.tex

\section{Appendix}

\begin{figure}[t]
	\includegraphics[width = 0.45\textwidth]{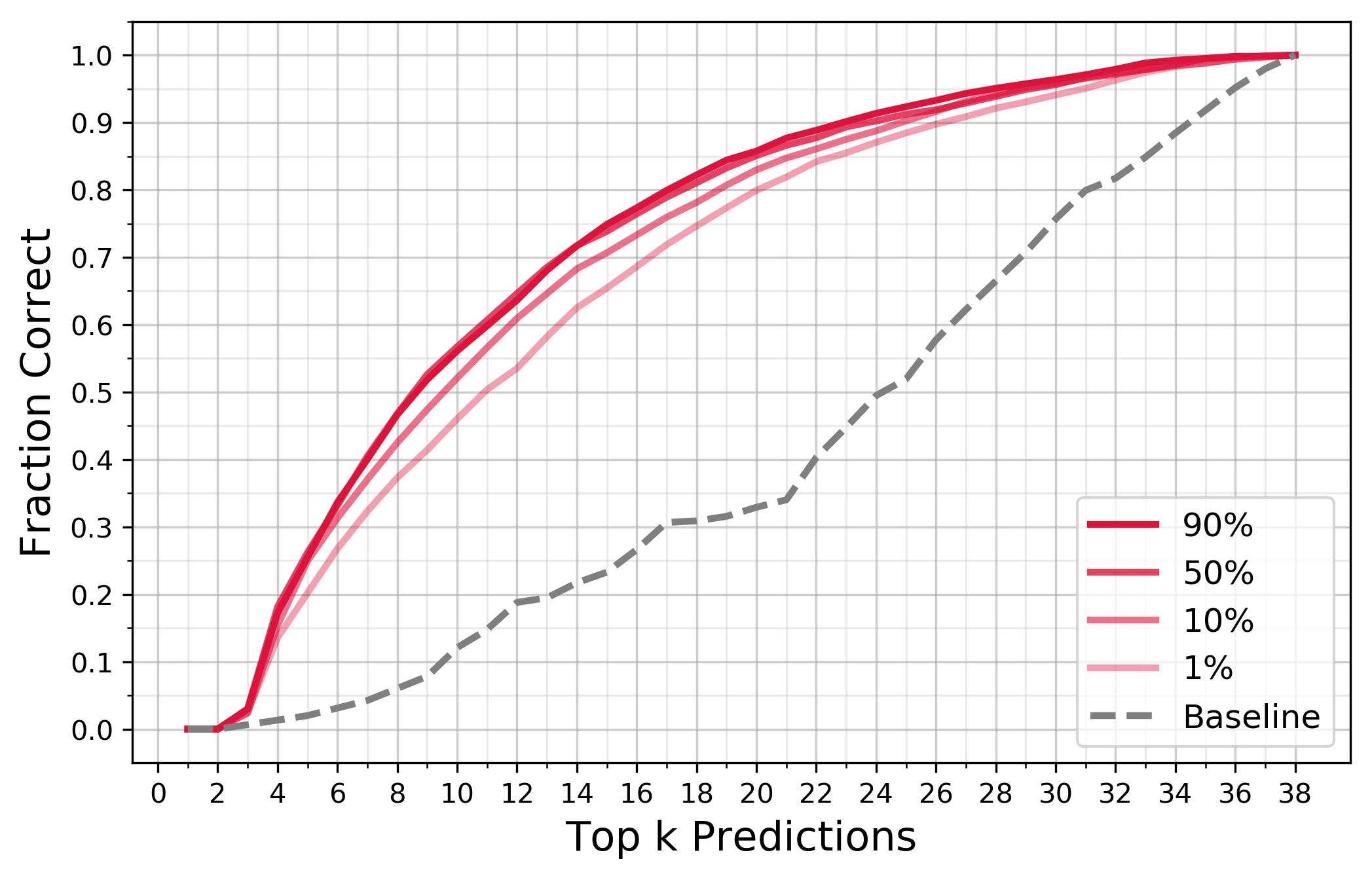}
	\caption{Here we vary the total number of unique programs in training sets and compare the resulting network performance on a fixed, semantically disjoint, test set. All sets were generated on length $\ell = 3$ programs with the non-uniform sampling method.}
	\label{fig:training-with-less}
\end{figure}

\subsection*{Specifics of neural network architectures considered}

The original DeepCoder network, summarized in detail in the supplemental material of \citep{balog2017deepcoder}, 
takes as input 5 I/O pairs, padded to a fixed length.
Each integer in the network inputs is sent through a trainable embedding layer, represented as 20-dimensional real-valued vectors; 
these embeddings are then concatenated together and passed through three fully-connected layers of size 256.
This yields 5 representations, one for each of the input/output pairs input into the network;
these are then averaged, and passed as input into a final sigmoidal activation layer, which outputs the predicted probabilities of the 34 components appearing in the program.
Note that the branching factor for a search tree (e.g. depth-first search) is larger than 34, since the lines which contain higher-order functions also require selection of one of the predicate functions; for lines which have two functions, the ranking order is determined by the smaller probability.

In the recurrent neural network, a long short-term memory (LSTM) network is added to produce a per-line heuristic.
Most of the architecture is unchanged: as in the original DeepCoder model, the inputs and outputs are sent through an embedding layer, concatenated, passed through three fully-connected layers, and then averaged across the five examples.
However, instead of predicting the probabilities of inclusion for each function directly,
this representation is instead provided as an input into the LSTM, which outputs a new representation for each line of the program.
As before, a final sigmoidal output layer emits probabilities for each of the 34 functions, but now it is applied across each line.
The embedding layer in this network is 50-dimensional, and the both the fully connected layers and the LSTM have 200 hidden units.
The result is a network which takes as arguments not just the input / output examples, but also a target ``number of lines'', and then returns estimates of probabilities that a function occurs on a per-line basis, rather than program-wide.

\subsection*{Influence of the size of the training set}

We investigated the effect of training on 90\%, 20\%, 10\% and 1\% of the possible programs. All the experiments here were done on programs of length $\ell = 3$, using non-uniform sampling to generate the training and test sets. Although we decrease the number of total unique programs in the training sets, the \emph{total} number of examples remain fixed for each set at $n=300000$. Although the later sets see less programs, they contain more examples per program. 

We observe in Figure~\ref{fig:training-with-less} that the performance on top-k prediction of the test set programs is not very sensitive to the amount of programs in the test set. In light of this fact, we choose to train with sets containing 10\% of the possible programs. This is good news, since in most settings, valid input-output pairs would be cheaper to generate than valid programs.  

\subsection*{Details of Restricted Domain data generation method}
\begin{figure*}[!htb]
	\centering
	\resizebox{0.95\textwidth}{!}{
		\fbox{
			\begin{minipage}{.3\textwidth}
				\textbf{Example program:}\\
				\texttt{a} $\gets$ \texttt{[int]}\\
				\texttt{b} $\gets$ \textsc{Filter} \texttt{(\%2==1)} \texttt{a}\\
				\texttt{c} $\gets$ \textsc{Map} \texttt{(+1)} \texttt{b}\\
				\texttt{d} $\gets$ \textsc{ScanL1} \texttt{(*)} \texttt{c}\\
			\end{minipage}%
			\begin{minipage}{.7\textwidth}
				\textbf{Example input, and incremental output:}\\
				\texttt{a = [-200, 144, 25, 66, -7, 38, -1, 14, 80, 81, 155]}\\
				\texttt{b = [25, -7, -1, 81, 155]}\\
				\texttt{c = [26, -6, 0, 82, 156]}\\
				\texttt{d = [26, -156, 0, 0, 0]}\\
			\end{minipage}
		}
	}
	\caption{A program with complex restrictions on inputs that will remain in the target range}
	\label{fig:DifficultExampleForPropBounds}
\end{figure*}

A value range for acceptable outputs is specified (initially $[-255, 256]$), as well as the maximum length of the output if it is a list (the length is chosen uniformly at random from one to ten). 
The program is evaluated backwards, computing for each intermediate variable a `safe'  range for its value that guarantees to have outputs in the target range. 

By applying this backward propagation of bounds to the whole program we find a suitable input range. 
Values are then sampled uniformly from this range to create the input(s), and the output is calculated by evaluating the program.  
If the resulting valid range for some input is empty or a singleton, then the program is discarded.

The short program described in Figure~\ref{fig:DifficultExampleForPropBounds} illustrates the approach and a key limitation of it. 
The function \textsc{ScanL1}\texttt{(*)} on list $A$ outputs a list whose value at position $k$ is the product $\Pi_{i=0}^k A[i]$.
For lists of length $5$ the input range for \textsc{ScanL1}\texttt{(*)} that guarantees outputs between $-256$ and $256$ is $\left[ -3, 3 \right] $.
Pushing this range back through \textsc{Map}\texttt{(+1)} gives a range for \texttt{b} of $\left[ -4, 2\right]$, which is unchanged by \textsc{Filter} \texttt{(\%2==1)}.
In fact any even number could be accepted as part of the input, and if the input contains $-1$ at some point then any value appearing subsequently in list can be large. 

\subsection*{Non-uniform Sampling}
\begin{figure}[tbp]
	\begin{center}
		\includegraphics[width=0.9\columnwidth]{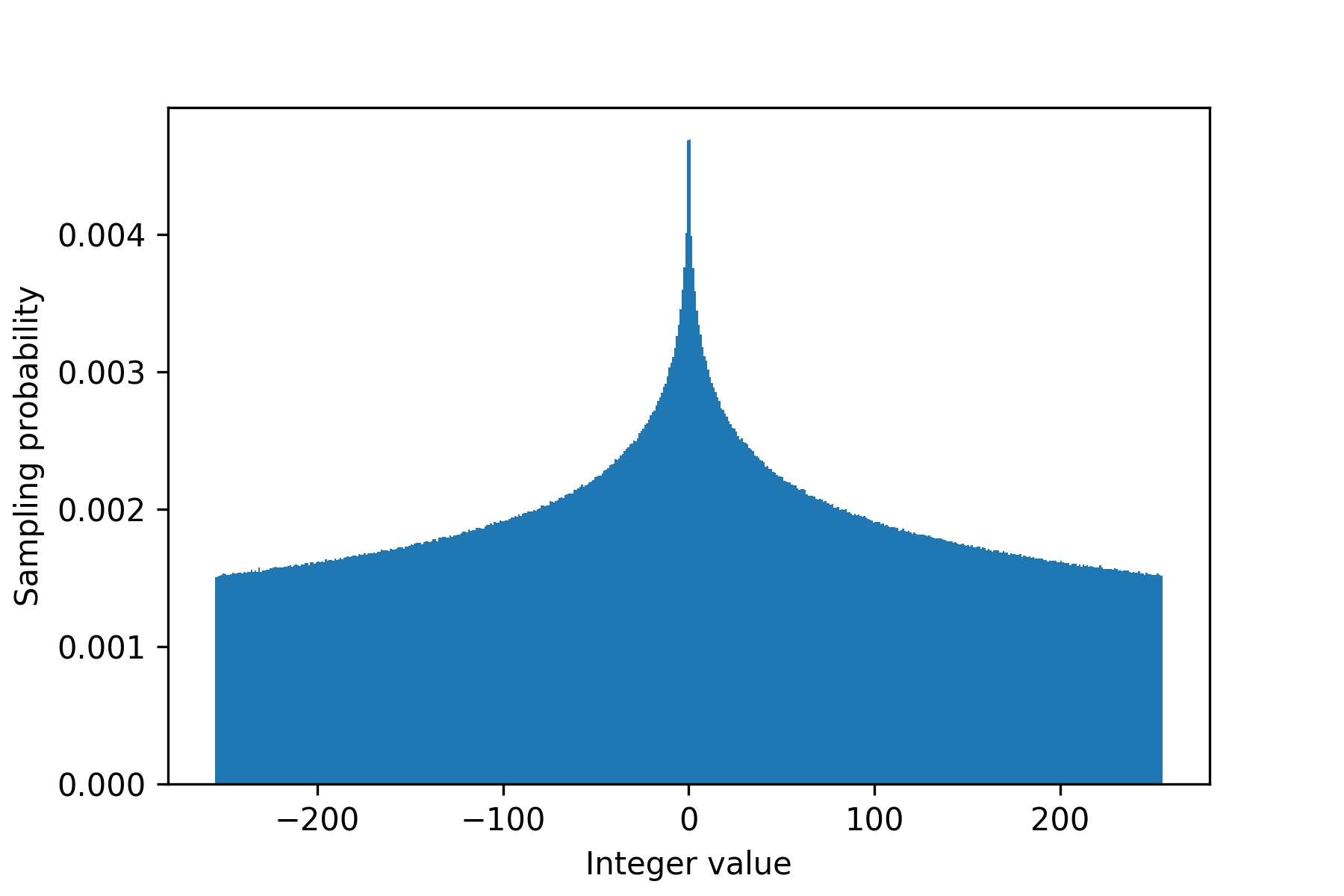}
		\caption{Marginal distribution over proposed integer values in the non-uniform sampling.}
		\label{fig:exp-sampling-dist}
	\end{center}
\end{figure}
The marginal distribution over sampled integer values is shown in Figure~\ref{fig:exp-sampling-dist}.

\paragraph{A note about empty outputs}
Some programs in the DSL output either null or the empty list on a large number of outputs. 
While these are informative to some extent, they dominated the examples generated for certain programs.
The process of propagating bounds through the program is focussed on the value ranges and does not naturally exclude empty outputs.
For example, it is not possible to specify a \emph{range} (other than one containing only a single element) for inputs to \textsc{Filter}\textsc{(\%2==0)} that guarantees a non-empty output. 
To ensure that empty outputs could not dominate we included post processing for both sampling methods which rejects empty outputs until $90\%$ of the permitted attempts have been made. 

\subsection*{Details of Simple Constraint Based Data Generation}
Our training data was made up of $25$ sets of $5$ examples for each program.
Due to the random choice of minimum input length there are $6$ versions of the initial SMT problem.
Subsequent SMT problems depend on the previous examples produced but because subsequent calls to the solver vary only slightly the examples generated can be very similar.
We experimented with randomly adding weak constraints such as setting the first element of an input to be odd or even or positive or negative in order to modify the problem slightly for each call to the solver, however for the experiments reported here the data was produced without any such random constraints and a few examples are indeed repeated across different sets.

\subsection*{The Effect of Neural Network Performance on Search}

\begin{figure}[t]
	\includegraphics[width = 0.45\textwidth]{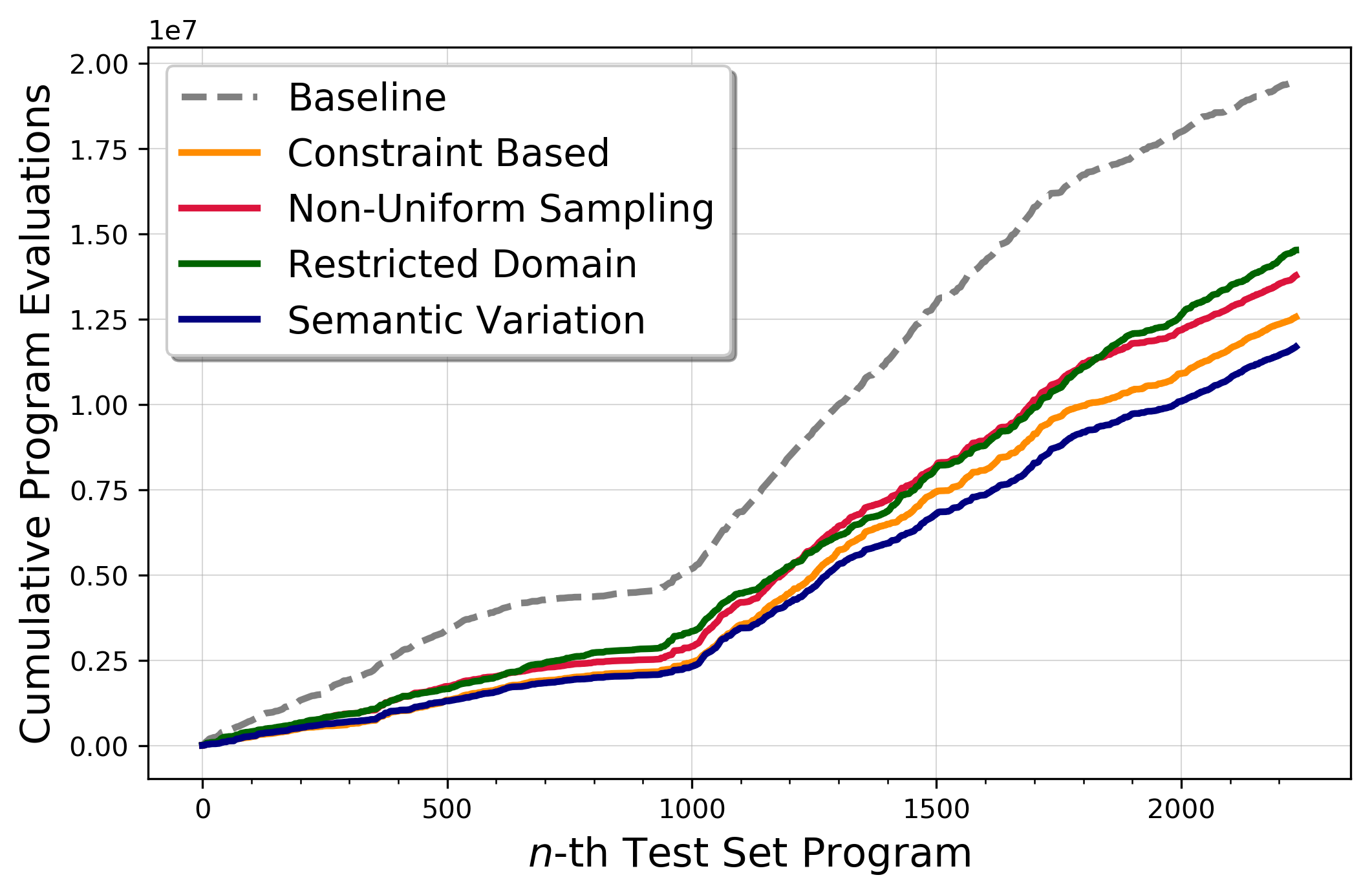}
	\caption{Relative improvement to a search procedure assisted by neural network predictions. The network was trained on the Semantic Variation data, we see how the loss in predictive performance is reflected in increased search effort}
	\label{fig:combined-test-set-search}
\end{figure}

Since the intention of training the neural network is to use it to aid a search procedure, we ran a depth first search based on the predictions made by each network.
The cumulative time taken  to complete the search by each network is shown in Figure~\ref{fig:combined-test-set-search}, showing the effect of reduced prediction accuracy on the time taken to find suitable programs.
The network trained on similar data saved around one quarter of the total search time across the test set over the worst performing network. 
This shows how the benefits of machine learning to programming by example may be overstated when only evaluated on ``friendly'' artificial data.

\end{document}